\documentclass[runningheads]{llncs}
\usepackage[square,sort,comma,numbers]{natbib}
\usepackage{lipsum}
\usepackage{algorithmic}
\usepackage{algorithm}
\usepackage{multicol}
\usepackage{times}
\usepackage{helvet}
\usepackage{courier}
\usepackage{hyperref}
\usepackage{amsmath}
\usepackage{amsfonts}
\usepackage{ntheorem}

\usepackage{multirow}
\usepackage{booktabs}
\usepackage{caption}
\usepackage{graphicx}
\usepackage{bbm}
\usepackage{float} 
\usepackage{subfigure}
\usepackage{subcaption}
\usepackage{xspace}

\usepackage{threeparttable}
\usepackage{xcolor}
\usepackage{colortbl}

\usepackage[T1]{fontenc}
\usepackage{graphicx}
\begin{document}
%
\title{Toward the Tradeoffs between Privacy, Fairness and Utility in Federated Learning}
\author{Kangkang Sun\inst{1} \and
Xiaojin Zhang\inst{2} \and
Xi Lin\inst{1} \and
Gaolei Li\inst{1} \and
Jing Wang \inst{1} \and
Jianhua Li\inst{1}}
\authorrunning{K. Sun et al.}

\institute{Shanghai Key Laboratory of Integrated Administration Technologies for Information Security, School of Electronic Information and Electrical Engineering, Shanghai Jiao Tong University, Shanghai, China\\
School of Computer Science and Technology, Huazhong University of Science and Technology, Wuhan, China\\
\email{\{szpsunkk, linxi234, gaolei\_li, wangjing08, lijh888\}@sjtu.edu.cn; xiaojinzhang@hust.edu.cn}}

\maketitle

\begin{abstract}
Federated Learning (FL) is a novel privacy-protection distributed machine learning paradigm that guarantees user privacy and prevents the risk of data leakage due to the advantage of the client's local training. Researchers have struggled to design fair FL systems that ensure fairness of results. However, the interplay between fairness and privacy has been less studied. Increasing the fairness of FL systems can have an impact on user privacy, while an increase in user privacy can affect fairness. In this work, on the client side, we use the fairness metrics, such as \textit{Demographic Parity} (DemP), \textit{Equalized Odds} (EOs), and \textit{Disparate Impact} (DI), to construct the local fair model. To protect the privacy of the client model, we propose a privacy-protection fairness FL method. The results show that the accuracy of the fair model with privacy increases because privacy breaks the constraints of the fairness metrics. In our experiments, we conclude the relationship between privacy, fairness and utility, and there is a tradeoff between these.

\end{abstract}

\section{Introduction}
Federated learning (FL) \citep{mcmahan2017communication,kairouz2021advances} is a novel distributed machine learning approach that guarantees user privacy by ensuring that user data does not leave the local area. However, FL has been plagued by two ethical issues: privacy and fairness \citep{chen2023privacy}. So far, most of the research has considered these two issues separately, but the existence of some kind of intrinsic equilibrium between the two remains unexplored. For example, privacy can come at the expense of model accuracy, however, for different groups of people training privacy results in different accuracies, with disadvantaged groups often suffering a greater cost in the training process. On the other hand, in order to ensure the fairness of the model and eliminate the bias in the training data or model \citep{agarwal2018reductions, berk2021fairness}, the client needs to share more data with the server, which seriously increases the user privacy risk. Therefore, it is an open issue to investigate the intrinsic connection between fairness and privacy in FL and to break the distress caused by its tradeoffs.

\textbf{Privacy Destroys Fairness}
The first observation is that the decrease in accuracy due to deep DP models has a disproportionately negative impact on underrepresented subgroups \citep{bagdasaryan2019differential}. DP-SGD enhances model “bias” in different distributions that need to be learned. Subsequently, in the study \citep{pujol2020fair}, the impact of DP on fairness in three real-world tasks involving sensitive public data. There are significant differences in the model outputs when stronger privacy protections are implemented or when the population is small. Many works \citep{tran2021differentially, esipova2022disparate} have attempted to find reasons why privacy destroys fairness.

\textbf{Fairness Increases Privacy Risk}
The client's dataset is usually unbalanced and biased. This bias is gradually amplified during the machine learning process. For example, when a model is trained for accuracy, the model's predictions will correlate with gender, age, skin, and race in a certain demographic group \citep{zafar2017fairness, berk2021fairness,chouldechova2017fair}. 

Privacy and fairness are two important concepts in FL, and violating either one is unacceptable. Therefore, this paper explores the intrinsic relationship between privacy and fairness in FL and designs a privacy protection method for fair federated learning, to improve the model learning performance while ensuring the privacy and fairness constraint. 

\textit{Relationship of fairness and privacy.} We attempt to explore the relationship between fairness and privacy in FL. Intuitively, there is some intrinsic connection between fairness and privacy, and some balance between fairness, privacy, and utility.

\begin{itemize}
    \item \textit{Fairness:} We consider three fairness metrics, including Demographic Parity (DemP), Equalized Odds (EO) and Disparate Impact (DI). Comparing the research \citep{pujol2020fair}, we design the optimization function to be more complex, taking into account privacy and fairness constraints.
    \item \textit{Privacy:} In this paper, we consider privacy-protection methods for fair Federated Learning based differential privacy.
\end{itemize}
Our contributions can be summarized as follows:
\begin{itemize}
    \item A privacy-protection fairness FL method is proposed, in order to protect the model privacy of the client while sharing model parameters. Our proposed method is mainly divided into two parts: fairness training and privacy-protection training. Specifically, the client first trains a fairness proxy model and then trains a privacy-protection model based on that proxy model.
    \item In this paper, We experimentally obtained that the increase in privacy destroys the fairness of the model but appropriately increases the accuracy of the model. In order to improve the accuracy of the model and to ensure the fairness of the model, we designed private fair algorithms \ref{al: Imp-LDP}.
    \item We demonstrate the superiority of our proposed method and algorithms based on \textit{Adult} datasets comparing popular benchmark \textit{FedAvg} algorithms. Experiments prove that our algorithm can effectively guarantee model privacy in fair FL.
\end{itemize}
\section{Related Work}
\subsection{Fairness of FL}
Fairness of FL is defined in two ways: client fairness \citep{li2019fair, martinez2020minimax, yu2020fairness, karimireddy2020scaffold} and algorithmic fairness \citep{hardt2016equality}. Algorithmic fairness has been extensively studied in traditional centralized machine learning through debiasing methods \citep{kairouz2021advances}. However, due to the fact that in FL, the server does not have access to client-side local data, it is already difficult to estimate the global data distribution simply by debiasing either server-side or client-side \citep{mcmahan2017communication}. Much research has focused on client fairness in FL, such as in augmenting client data aspect \citep{hao2021towards, jeong2018communication}, in the client data distribution aspect \citep{duan2020self, wang2020optimizing}. From a model perspective, training a separate fairness model for each client is an open problem.

\subsection{Privacy of FL}
Many recent studies have focused on FL privacy risks \citep{gehlhar2023safefl, lowy2023private, shao2023survey, bietti2022personalization}. A diversity of privacy-protection techniques have been proposed to discourage the risk of privacy leakage for users, including cryptographic techniques and the perturbation approach \citep{chen2023privacy}. Cryptographic approaches allow computation on encrypted data and provide strict privacy guarantees. However, they are computationally expensive compared to non-encryption methods \citep{xu2021privacy}. This computational overhead seriously affects the machine learning training process, especially with a large number of parameters in the model. Therefore, the current state-of-the-art privacy-protection methods are perturbation-based, such as the DP mechanism \citep{geyer2017differentially, wei2020federated, wu2021fedcg,scheliga2022precode}. The shuffler model is proposed to amplify the privacy of LDP's poor performance in comparison with the central DP mechanisms \citep{raskhodnikova2008can, erlingsson2019amplification, cheu2019distributed, balle2020private, ghazi2021power, girgis2021shuffled}. Most research based on Shuffler's model has focused on the study of tradeoffs between privacy, utility, and communication \citep{chen2022fundamental, girgis2021shuffled, li2023privacy, zhou2022multi, balle2019privacy}.  However, there is very little research on the privacy protection of fair federated learning.

\subsection{ Fairness and Privacy of FL}
Recently, some work \citep{chen2023privacy, pujol2020fair} has led to inconsistent reductions in accuracy due to private mechanisms for classification \citep{farrand2020neither} and generation tasks \citep{ganev2022robin}. Because of the tension between fairness and privacy, researchers often need to make trade-offs between the two perceptions \citep{bagdasaryan2019differential, esipova2022disparate, tran2021differentially}. The trade-off may be to increase privacy preservation at the expense of fairness, i.e., by adopting a loose notion of fairness rather than a precise one or vice versa \citep{berk2021fairness, chouldechova2017fair}.
\begin{table}[ht]
\centering
\caption{Private and Fair Federated Learning}
\begin{threeparttable}
\resizebox{\linewidth}{!}{
\begin{tabular}{cccccc} 
\toprule
\multirow{2}{*}{References} & \multirow{2}{*}{\begin{tabular}[c]{@{}c@{}}Privacy \\Metrics\end{tabular}} & \multirow{2}{*}{\begin{tabular}[c]{@{}c@{}}Fairness~\\Metrics\end{tabular}} & \multicolumn{2}{c}{Techniques}                                                                                                                   & \multirow{2}{*}{Trade-off type}  \\ 
\cline{4-5}
                           &                                                                            &                                                                             & Privacy                                                                         & Fairness                                                       &                                 \\ 
\hline
\citep{lamy2019noise}                        & $\epsilon$-DP                                                                       & EOs \& DemP                                                                 & \begin{tabular}[c]{@{}c@{}}Class conditional\\noise\end{tabular}                & \begin{tabular}[c]{@{}c@{}}Fairness \\constraints\end{tabular} & I                               \\ 
\hline
\citep{jagielski2019differentially}                         & $(\epsilon, \delta)$-DP                                                                   & EOs                                                                         & \begin{tabular}[c]{@{}c@{}}Exponential mechanism \&\\Laplace noise\end{tabular} & \begin{tabular}[c]{@{}c@{}}Fairness \\constraints\end{tabular} & /                               \\ 
\hline
\citep{lowy2023stochastic}                        & $(\epsilon, \delta)$-DP                                                                   & EOs \& DemP                                                                   & DP-SGDA                                                                         & \begin{tabular}[c]{@{}c@{}}ERMI \\regularizer\end{tabular}     & II                              \\ 
\hline
\citep{tran2021differentially}                        &    \begin{tabular}[c]{@{}c@{}}$(\alpha, \epsilon_p)$-Renyi \\DP \end{tabular}                                                         & EOs, AP \& DemP                                                             & DP-SGD                                                                          & \begin{tabular}[c]{@{}c@{}}Fairness \\constraints\end{tabular} & II                              \\ 
\hline
\citep{kilbertus2018blind}                        & /                                                                          & EA                                                                          & MPC                                                                             & \begin{tabular}[c]{@{}c@{}}Fairness\\constraints\end{tabular}  & II                              \\ 
\hline
\citep{diana2022multiaccurate}                         & /                                                                          & EOs                                                                         & \begin{tabular}[c]{@{}c@{}}Proxy \\attribute\end{tabular}                       & Post-processing                                                & II                              \\ 
\hline
\citep{wang2020robust}                        & /                                                                          & DemP                                                                        & \begin{tabular}[c]{@{}c@{}}Noisy \\attribute\end{tabular}                       & \begin{tabular}[c]{@{}c@{}}Fairness \\constraints\end{tabular} & II                              \\ 
\hline
\citep{awasthi2020equalized}                          & /                                                                          & EOs                                                                         & \begin{tabular}[c]{@{}c@{}}Noisy \\attribute\end{tabular}                       & Post-processing                                                & II                              \\ 
\hline
\textbf{Our Method}                & $(\epsilon, \delta)$-DP                                                                   & EOs, DemP,  DI                                                              & Gaussian Noise                                                                  & \begin{tabular}[c]{@{}c@{}}Fairness\\constraints\end{tabular}  & II                              \\
\bottomrule
\end{tabular}
}
      \begin{tablenotes} 
		\item I: Trade fairness for privacy.  II: Trade privacy for fairness.
            \item EOs: Equalized Odds. DemP: Demographic Parity. AP: Accuracy Parity. EA: Equal Accuracy. 
            \item DI: Disparate Impact.
     \end{tablenotes} 
\end{threeparttable}
\end{table}
\section{Preliminaries}
\subsection{Fairness in FL}
We consider the following fairness metrics, including DemP, EO and DI. DemP denotes the same probability of getting a chance under some sensitive attribute. EO is a subset of DP, defined as the probability of getting a chance on a given aspect is the same for different sensitive attributes. Let $X, Y$ be the sensitive attribute and the true label, respectively.
For example, $Y=1$ often represents the condition of being able to apply for a loan, and $Y=0$ is the condition of not meeting the loan. Thus, on the opportunity to apply for a loan, the output has the same probability for each person (characteristic), and then this is EO fairness.
\begin{definition} { \rm \textbf{(Demographic Parity (DemP))}} \citep{hardt2016equality} We say that a predictor $f$ satisfies demographic parity with respect to attribute $A$, instance space $X$ and output space $Y$, if the output of the prediction $f(X)$  is independent of the sensitive attribute $\mathcal{A}$. For $\forall a \in A$ and $p \in\{0,1\}$:

\begin{equation}
\mathbf{P}[f(X)=p \mid \mathcal{A}=a]=\mathbf{P}[f(X)=p]
\end{equation}
Given $p \in \{0,1\}$, for $\forall a \in A$:

\begin{equation}
\mathbb{E}[f(X) \mid \mathcal{A}=a]=\mathbb{E}[f(X)]
\end{equation}
However, the left and right terms of the above equality are often not the same. 
Then, the loss $l_{DemP}$ of DemP can be defined as follows:

\begin{equation}
    l_{DemP} = \mathbb{E}[f(X) \mid \mathcal{A}=a] - \mathbb{E}[f(X)] 
\end{equation}
\end{definition}
\begin{definition}{ \rm \textbf{(Equalized Odds (EO))} }\citep{hardt2016equality} We say that a predictor $f$ satisfies equalized odds with respect to attribute $A$, instance space $X$ and output space $Y$, if the output of the prediction $f(X)$  is independent of the sensitive attribute $\mathcal{A}$ with the label $\mathcal{Y}$. For $\forall a \in \mathcal{A}$ and $p \in\{0,1\}$:
\begin{equation}
\mathbf{P}[f(X)=p \mid \mathcal{A}=a, Y=y]=\mathbf{P}[f(X)=p \mid Y=y]
\end{equation}
Given $p \in \{0,1\}$, for $\forall a \in A, y \in Y$:
\begin{equation}
\mathbb{E}[f(X) \mid \mathcal{A}=a, Y=y]=\mathbb{E}[f(X) \mid Y=y]
\end{equation}
Then, the loss $l_{EO}$ of EO can be defined as follows:
\begin{equation}
    l_{EO} = \mathbb{E}[f(X) \mid \mathcal{A}=a, Y=y] - \mathbb{E}[f(X) \mid Y=y]
\end{equation}
\end{definition}
\begin{remark}
    A binary predictor $f$, satisfying the demographic parity, is a special instance of equalized odds.
\end{remark}

\begin{definition} {\rm \textbf{(Disparate Impact (DI))}} \citep{pujol2020fair}
We say that a predictor $f$ satisfies disparate impact with respect to attribute $\mathcal{A}$, if the output of the prediction $f(X)$ is independent of the sensitive attribute $\mathcal{A}$ with a similar proportion of the different groups. For $a \in \{0,1\}$, we have:

\begin{equation}
\min \left(\frac{\mathbf{P}(f(x)>0 \mid a=1)}{\mathbf{P}(f(x)>0 \mid a=0)}, \frac{\mathbf{P}(f(x)>0 \mid a=0)}{\mathbf{P}(f(x)>0 \mid a=1)}\right)=1
\end{equation}
For $i \in [0,n]$ and $i$ is a positive integer:
{\small \begin{equation}
\min \left(\frac{\mathbf{P}(f(x)>0 \mid a=i+1)}{\mathbf{P}(f(x)>0 \mid a=i)}, \frac{\mathbf{P}(f(x)>0 \mid a=0)}{\mathbf{P}(f(x)>0 \mid a=n)}\right)_{i = 0}^{n}=1
\end{equation}}
\end{definition}
Then, the loss $l_{DI}$ of DI can be defined as follows:
{\footnotesize \begin{equation}
    l_{DI} = \min \left(\frac{\mathbf{P}(f(x)>0 \mid a=i+1)}{\mathbf{P}(f(x)>0 \mid a=i)}, \frac{\mathbf{P}(f(x)>0 \mid a=0)}{\mathbf{P}(f(x)>0 \mid a=n)}\right)_{i = 0}^{n} - 1
\end{equation}}

\subsection{Privacy in FL}
The local dataset of clients contains sensitive data, which requires protecting the sensitive attributes while training. Differential Privacy (DP) is a privacy protection technique designed to safeguard individual data while allowing data analysis and mining \citep{dwork2014algorithmic}. Local Differential Privacy (LDP) is deployed on clients to protect the attributes of the local dataset, in order to make sure that any algorithm built on this dataset is differentially private. The $\epsilon$- differentially private mechanism $\mathcal{M}$ is defined as follows:
\begin{definition} {\rm \textbf{(Local Differential Privacy (LDP))} }\citep{dwork2014algorithmic}
    A randomize algorithm $\mathcal{M}: X \to Y$ satisfies $(\epsilon,\delta)$-LDP with respect to a input set $X$ and a noise output set $Y$, if $\forall x, x^{'} \in X$ and $\forall y \in Y$ hold:
\begin{equation}
    \mathbf{P}[\mathcal{M}(x)=y] \leq e^{\epsilon}  \mathbf{P}\left[\mathcal{M}\left(x^{\prime}\right)=y\right]+\delta
\end{equation}
\end{definition}

\begin{definition}[Gaussian Mechanism] Assume that a deterministic function $f:\mathcal{M} X \to Y$ with $\Delta_2(f)$ sensitivity, then for $\forall \delta \in (0, 1)$, random noise follows a normal distribution $\mathcal{N}(0, \sigma^2)$, the mechanism $\mathcal{M}(d) = f(d)+\mathcal{N}(0, \sigma^2)$ is  $(\epsilon, \delta)$-DP, where

\begin{equation}
    \epsilon \geq \frac{\sqrt{2 \ln (1.25 / \delta)}}{\frac{\sigma}{\Delta_2 f}}, \quad \Delta_2(f) = \max _{d, d^{\prime} \in \mathcal{D}}\left\|f(d)-f\left(d^{\prime}\right)\right\|_2
\end{equation}
\label{de: gu}
\end{definition}
\subsection{Problem Formulation}

There is a set of $n$ clients in the FL system, where $m \in n$ clients are selected to participate in the FL training process. The clients have its own local dataset $\mathcal{D}_i = \{d_1,...,d_n\}$. Let $\mathcal{D}=\bigcup_{i=1}^n \mathcal{D}_i$ denote the entire dataset and $f(\theta_i, d_i)$ as the loss function of client $i$, where the parameter $\theta \in \Theta$ is the model parameter. There are $m \in n$ clients
The clients are connected to an untrusted server in order to solve the ERM problem $F_i\left(\theta, \mathcal{D}_i\right)=\frac{1}{b} \sum_{j=1}^b f\left(\theta, d_{i j}\right)$, where local estimated loss function dependent on the local dataset $\mathcal{D}_i$, and $b$ is the local batch size. We give the ERM problem \citep{kairouz2021advances} in FL, as follows:

\begin{equation}
\begin{aligned}
 \arg \min _{\theta \in \mathcal{C}} & \left(F(\theta):=\frac{1}{m} \sum_{i=1}^m F_i(\theta)\right), \\
    s.t.&  \quad   l_{DemP} < \mu_{DemP}, \\
         &  \quad  l_{EO} < \mu_{EO}, \\
         &  \quad  l_{DI} < \mu_{DI},
\end{aligned}
\label{eq: ERM_f}
\end{equation}
where the $l_{DemP}, l_{EO}, l_{DI}$ are the loss constraint of DemP, EO and DI, respectively.
We use the Lagrangian multiplier \citep{pujol2020fair} to transform the ERM problem (\ref{eq: ERM_f}) into a Min-Max problem, as follows:

\begin{equation}
    \begin{aligned}
    F(\theta, \lambda, l) &= \arg \min_{\theta_i \in \Theta} \max_{\lambda_{ij} \in \Lambda} \frac{1}{m}\sum_{i=1}^m \left \{\frac{1}{b} \sum_{j=1}^b f_i (\theta_i + d_{ij}) + \lambda_{ij} l_k\right\},  \\
   & k \in \{DemP, EO, DI\}, 
    \end{aligned}
    \label{eq: Local}
\end{equation}
where the parameter $\lambda \in \Lambda$ is the Lagrangian multiplier. 
In this fairness stage, the purpose is to train the proxy model under the fairness matrixes, which is to solve the optimization problem.  For the optimization problem (\ref{eq: Local}), there is the Lagrangian duality between the following two functions:
\begin{equation}
    \begin{aligned}
        \min_{\theta \in \Theta} \quad \max_{\lambda \in \Lambda} F(\theta, \lambda, l), \\
        \max_{\lambda \in \Lambda} \quad \min_{\theta \in \Theta}   F(\theta, \lambda, l).
    \end{aligned}  
    \label{eq: dual}
\end{equation}
In order to solve the above dual optimization problem (\ref{eq: dual}), many works assume the function $F$ is Liptches and convex and obtain a $\nu$\textit{-approximate saddle point} of Langrangian, with a pair ($\widehat{\theta}, \widehat{\lambda}$), where

\begin{equation}
    \begin{aligned}
        &F(\widehat{\theta}, \widehat{\lambda}, l) \leq F(\theta, \widehat{\lambda}, l) + \nu \quad \text{for all} \quad \theta \in \Theta,   \\
        &F(\widehat{\theta}, \widehat{\lambda}, l) \geq F(\widehat{\theta}, \lambda, l) - \nu \quad \text{for all} \quad  \lambda \in \Lambda.
    \end{aligned}
\end{equation}
Therefore we can get the Max-Min and the Min-Max dual problems are equivalent in the ERM problem (\ref{eq: ERM_f}). In order to search for the optimal value ($\theta^*, \lambda^*$) (or \textit{Nash Equilibrium} in-game) of the problem (\ref{eq: ERM_f}), many works study the fairness model by many approaches, such as the Zero-Game \cite{jagielski2019differentially, mozannar2020fair}, Distributionally Robust Optimization (DRO) \cite{wang2020robust}, and Soft Group Assignments \cite{wang2020robust}. In this paper, the fair model is optimized by the DRO method through a Lagrangian dual multiplier in clients, and the model parameters are then transmitted to the server for model aggregation through privacy-protection. 
\section{Method}
In this section, we design privacy protection for fair federated learning based on differential privacy. In section 4.1, the fair model in the FL system is obtained by the Algorithm \ref{al: fair}, where the fair model of each client can be optimized under constraints of \textit{DemP}, \textit{EO} and \textit{DI}. In section 4.2, we design a privacy protection algorithm \ref{al: Imp-LDP} for the fair model optimized in section 4.1.
\subsection{Fairness Predictor (Model) in Client}
Firstly, the clients train their own personalized fairness predictor, and we designed an Algorithm \ref{al: fair} to train the fair model on each client. In the Algorithm \ref{al: fair} line 5 and line 7, the optimal values ($\theta^*, \lambda^*$) are derived from the partial differential expression of the ERM problem (\ref{eq: ERM_f}). Secondly, each $\theta_i$ and $\lambda_i$ update their own information according to the partial differential expression in Algorithm \ref{al: fair} line 6 and line 8. Finally, after time $T_1$ rounds, the fair model of the client $i$ is output.
\begin{algorithm}
    \caption{\texttt{Fair-SGD} for client}
    \renewcommand{\algorithmicrequire}{\textbf{Input:}}
\renewcommand{\algorithmicensure}{\textbf{Output:}}
    \begin{algorithmic}[1]
    \label{al: fair_stage}
    \REQUIRE Local loss function $f(\cdot)$, train dataset $\mathcal{D}_i$, learning rate $\eta$, batch size $B$
    \STATE Initialize: $f_i (\theta_i) \leftarrow$ random, $\lambda_i \leftarrow$ max value
    \FOR{Each client $i \in \mathcal{N}$}
        \FOR{$t \in T_1$}
            \STATE Take a random batch size $B$ and $j \in B$  
            \STATE For $\theta_i$: $\mathbf{g}_t\left(x_j\right) \leftarrow \nabla_{\theta_{(i, t)}} f_i(\cdot)$
            \STATE $\theta_{(i, t+1)} \leftarrow \theta_{(i, t)}-\eta_t \mathbf{g}_t\left(x_j\right) $
            \STATE For $\lambda_i$, $\mathbf{g}_t^{\prime}\left(x_j\right) \leftarrow \nabla_{\lambda_{(i, t)}} f_i(\cdot)$
            \STATE $\lambda_{(i, t+1)} \leftarrow \lambda_{(i, t)}+\eta g_t^{\prime}\left(x_j\right)$ 
        \ENDFOR
    \ENDFOR
        \ENSURE Fair model $f_i(\theta_i)$
    \end{algorithmic}
    \label{al: fair}
\end{algorithm}
\subsection{Privacy Protection Method in Fair FL}
In this section, we design a privacy-protection fairness FL framework to protect the privacy and fairness of sensitive datasets in clients. As the above section, there is a trade-off between privacy, fairness and accuracy in the FL system. In this paper,  we designed a privacy-protection algorithm, named FedLDP Algorithm \ref{al: Imp-LDP}, based on the FedAvg algorithm.

\textbf{FedLDP:} In the algorithm, we design to add differential privacy preservation to the fairness model training process in algorithm \ref{al: Imp-LDP}. The algorithm, while reducing privacy consumption, can effectively improve the training accuracy of the model. Moreover, the algorithm does not guarantee that the intermediate entities are trustworthy, so the shuffler model is hijacked or attacked without any impact on user privacy.

\begin{algorithm}
    \caption{\texttt{FedLDP}}
    \renewcommand{\algorithmicrequire}{\textbf{Input:}}
    \renewcommand{\algorithmicensure}{\textbf{Output:}}
    \begin{algorithmic}[1]
    \label{al: Imp-LDP}
\REQUIRE Selected clients $m$, the local dataset $\mathcal{D}_i$ of client \ $i$, Maximum $L_2$ norm bound $C$, local privacy budget $\varepsilon_l$
        \STATE Initial the local model and download the global gradients from the server
        \FOR{$i \in m$ in parallel}
            \STATE \texttt{Fairness stage} in Algorithm (\ref{al: fair_stage})
            \STATE $\mathbf{g}_t\left(x_j\right) \leftarrow \nabla_{\theta_{(i, t)}} f_i(\cdot)$
           \STATE $\overline{\mathbf{g}}_t\left(x_j\right) \leftarrow \mathbf{g}_t\left(x_j\right) / \max \left(1, \frac{\left\|\mathbf{g}_t\left(x_j\right)\right\|_2}{C}\right)$
           \STATE $\tilde{\mathbf{g}}_t\left(x_j\right) \leftarrow \frac{1}{B}\left(\sum_i \overline{\mathbf{g}}_t\left(x_j\right)+\mathcal{N}\left(0, \sigma^2 C^2 \mathbf{I}\right)\right)$

            \STATE $\theta_{(i, t+1)} \leftarrow \theta_{(i, t)}-\eta_t \tilde{\mathbf{g}}_t\left(x_j\right)$
        \ENDFOR
        \STATE \quad \quad \textbf{Server}
        \STATE \textbf{Aggregate:} $\overline{\mathbf{g}}_t \leftarrow \frac{1}{\mathcal{N}_t} \sum_{i \in \mathcal{N}_t} \boldsymbol{w}_t\left(d_{i j}\right)$
        \STATE \textbf{Gradient Descent:} $\theta_{t+1}^G \leftarrow \theta_{t}^{G} + \mathbf{g}_t$
    \end{algorithmic}
\end{algorithm}

\section{Experiments}

\subsection{Dataset and Experimental Settings} 
In order to test the performance proposed in this paper, we use the \textit{Adult} \citep{Adult}, which is extracted from the U.S. Census dataset database, which contains 48,842 records, with 23.93\% of the annual income greater than \$50k and 76.07\% of the annual income less than \$50k, and has been divided into 32,561 training data and 16,281 test data. The class variable of this dataset is whether the annual income is more than \$50k or not, and the attribute variables include 14 categories of important information such as age, type of work, education, occupation, etc., of which 8 categories belong to the category discrete variables and the other 6 categories belong to the numerical continuous variables. This dataset is a categorical dataset that is used to predict whether or not annual income exceeds \$50k. We choose race as the sensitive attribute, including white person and black person.

\subsection{Experimental Hyperparameter Settings}
In the experiment, each client applied three (100$\times$100) fully connected layers.
\subsubsection{Machines}
The experiment was run on an ubuntu 2022.04 system with an intel i9 12900K CPU, GeForce RTX 3090 Ti GPU, and pytorch 1.12.0, torchversion 0.13.0, python 3.8.13.
\subsubsection{Software}
We implement all code in  \href{https://pytorch.org/}{PyTorch} and the \href{https://github. com/litian96/fair\_flearn}{fair\_learn} tool. 
\subsection{Performance Comparison Results}
\begin{figure}[ht]
    \centering
    \begin{minipage}{0.4\linewidth}
        \centering
        \centerline{
        \includegraphics[width=1\textwidth]{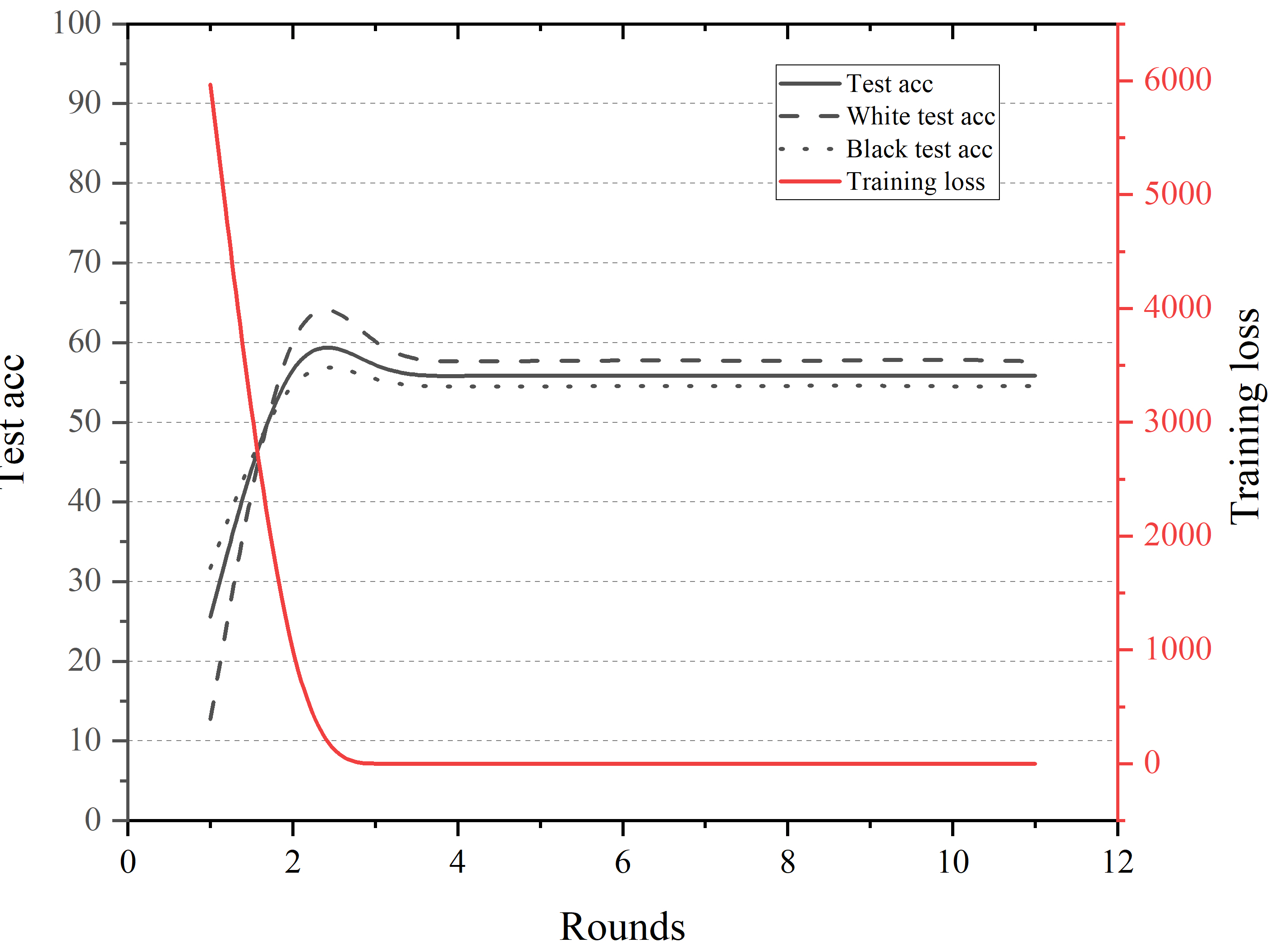}
        }
        \centerline{(\textit{a}) Fairness predictor with no privacy}
    \end{minipage}
    \begin{minipage}{0.4\linewidth}
        \centering
        \centerline{
        \includegraphics[width=1\textwidth]{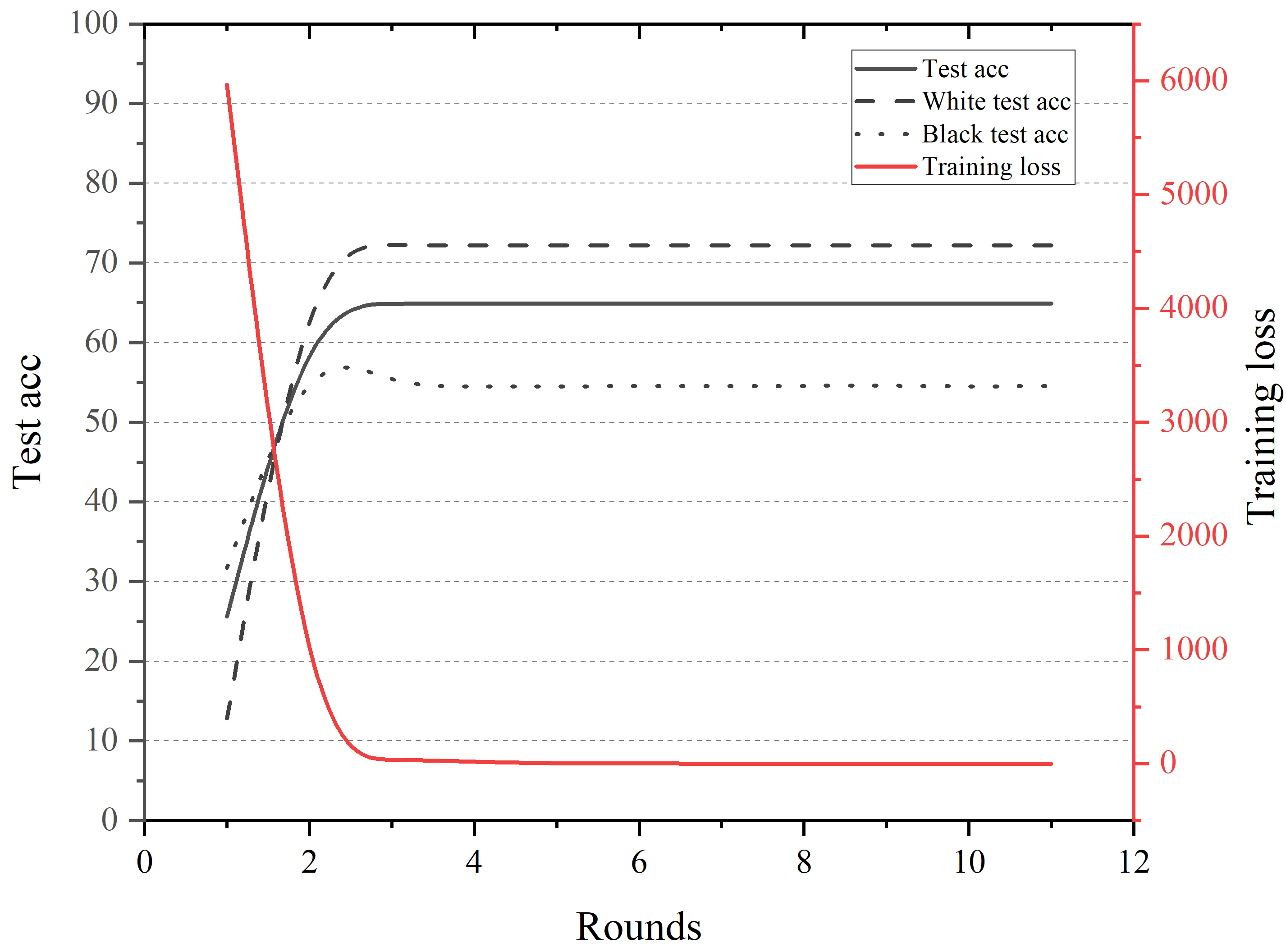}
        }
        \centerline{(\textit{b}) Fairness predictor with privacy ($\mathcal{N}(0, 1)$)}
    \end{minipage}
    \caption{The average test accuracy of the fair stage training process in FL settings with 5 clients on \textit{Adult} dataset. (\textit{a}) and (\textit{b}) are the training results with no privacy and privacy ($\mathcal{N}(0, 1)$), respectively. (\textit{a}) is shown that the test accuracy of sensitive data \textit{black} and \textit{white} are approximately the same for both. With the addition of noise privacy, test accuracy improves but fairness decreases, shown in (\textit{b}).} 
    
    \label{fig: fairstage}
\end{figure}

In the experiment, we compared the test accuracy between different algorithms. In the FL system, we tested both cases of fairness training without noise, and fairness training with noise, shown in Fig. \ref{fig: fairstage} (\textit{a}) and (\textit{b}). 
In Fig. \ref{fig: fairstage} (\textit{a}), the test accuracy of the white person is the same as the black person without noise in the client training process, while the fair client model with noise increases discrimination against different races in Fig. \ref{fig: fairstage} (\textit{b}).

Table \ref{tab: no_privacy} and Table \ref{tab:privacy} represent the test accuracy of differential clients in the FL system without noise and with noise, respectively. It can see from the table, that adding privacy improves the test accuracy for clients. The increase in privacy affects fairness because the increase in noise facilitates the optimizer to solve the global objective optimum while weakening the limitations of the fairness metrics, i.e., the constraints function $\lambda_{ij} l_k$.

\begin{table}[!h]
    \centering
    \begin{tabular}{|c|c|c|c|c|c|}
    \hline
         & Client $1$ & Client $2$ & Client $3$ & Client $4$ & Client $5$ \\
        \hline
    Black & 32.20 \% & 69.42 \% & 68.80\% & 68.96\% & 33.36 \%  \\
    \hline
    White & 12.26\% & 88.39 \% & 87.20\% & 87.05\% & 13.85 \%  \\
         \hline
    \end{tabular}
    \caption{The fair stage training process in FL settings with 5 clients (no privacy) on \textit{Adult} dataset.}
    \label{tab: no_privacy}
\end{table}
\begin{table}
    \centering
    \begin{tabular}{|c|c|c|c|c|c|}
    \hline
         & Client $1$ & Client $2$ & Client $3$ & Client $4$ & Client $5$ \\
        \hline
    Black & 66.63 \% & 73.75 \% & 68.96 \% & 69.41 \% & 67.79 \%  \\
    \hline
    White & 86.11 \% & 85.70 \% & 87.05\% & 88.39\% & 87.73 \%  \\
         \hline
    \end{tabular}
    \caption{The fair stage training process in FL settings with 5 clients (privacy $\mathcal{N}(0, 1)$) on \textit{Adult} dataset.}
    \label{tab:privacy}
\end{table}

\begin{figure}[H]
    \centering
        \begin{minipage}{0.4\linewidth}
        \centering
        \centerline{\includegraphics[width=0.8\textwidth]{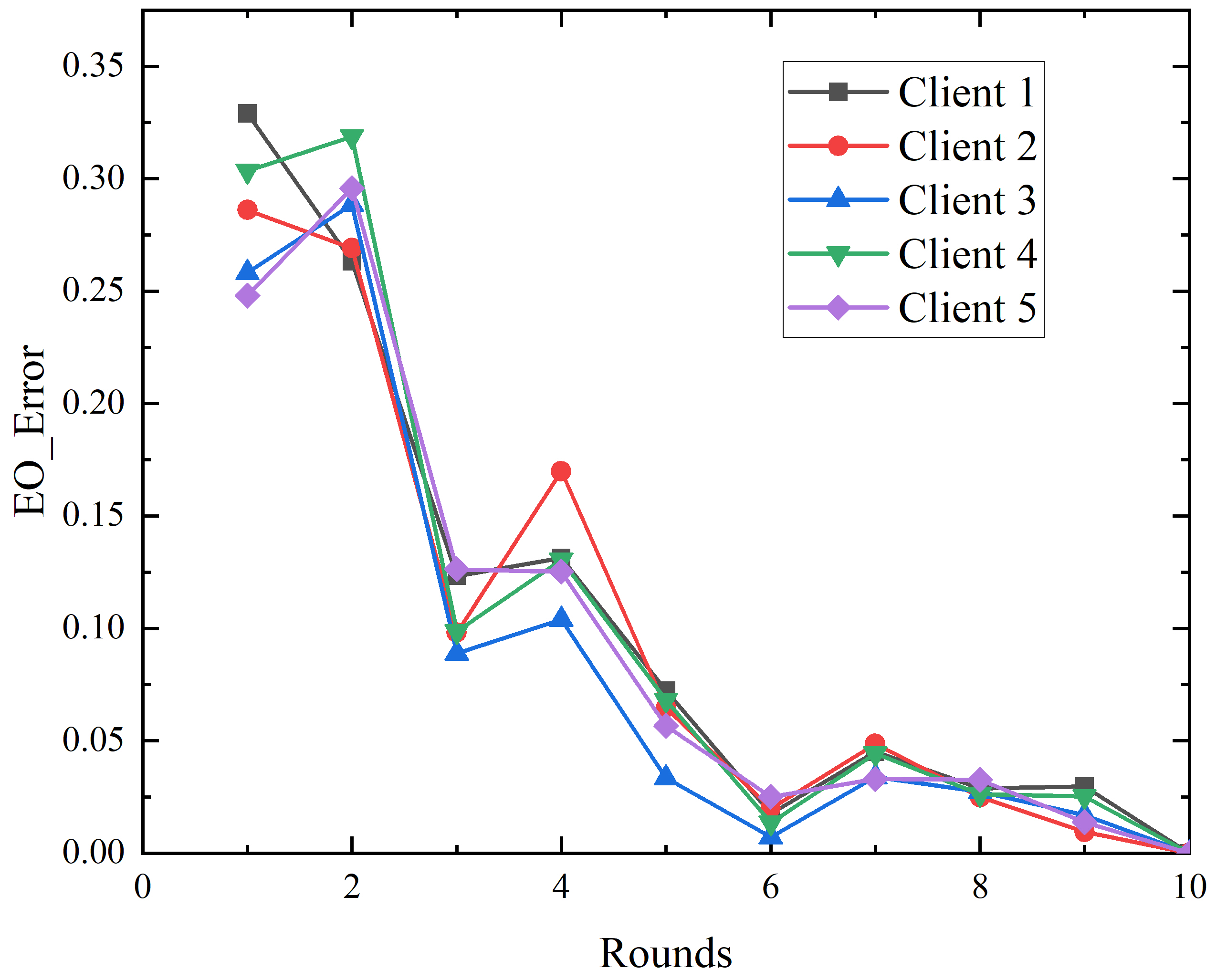}}
        \centerline{(\textit{a}) EO error without privacy}
        \end{minipage}
        \begin{minipage}{0.4\linewidth}
        \centering
        \centerline{\includegraphics[width=0.8\textwidth]{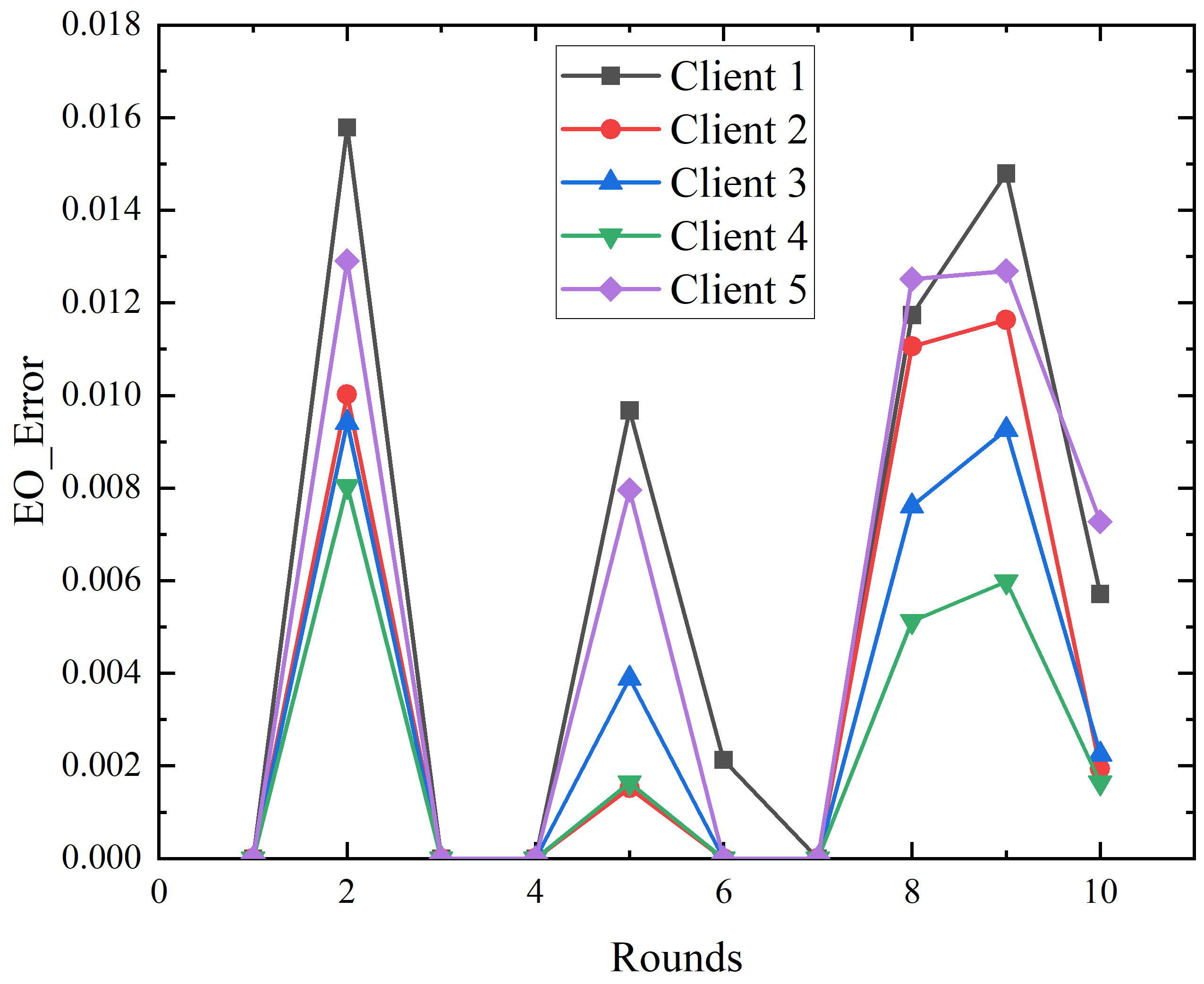}}
        \centerline{(\textit{b}) EO error with privacy}
        \end{minipage}
        \begin{minipage}{0.4\linewidth}
        \centering
        \centerline{\includegraphics[width=0.8\textwidth]{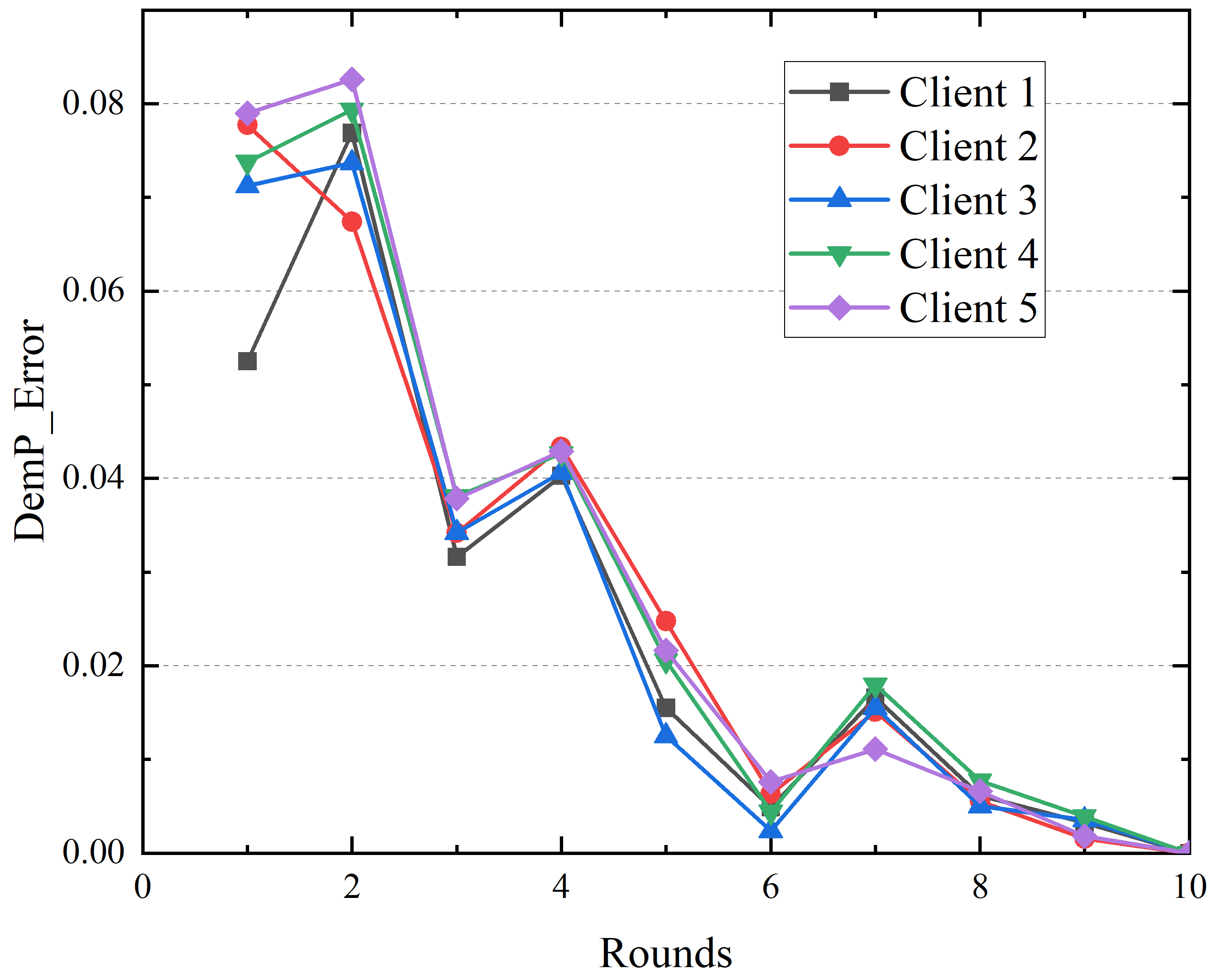}}
        \centerline{(\textit{c}) DemP error without privacy}
        \end{minipage}
        \begin{minipage}{0.4\linewidth}
        \centering
        \centerline{\includegraphics[width=0.8\textwidth]{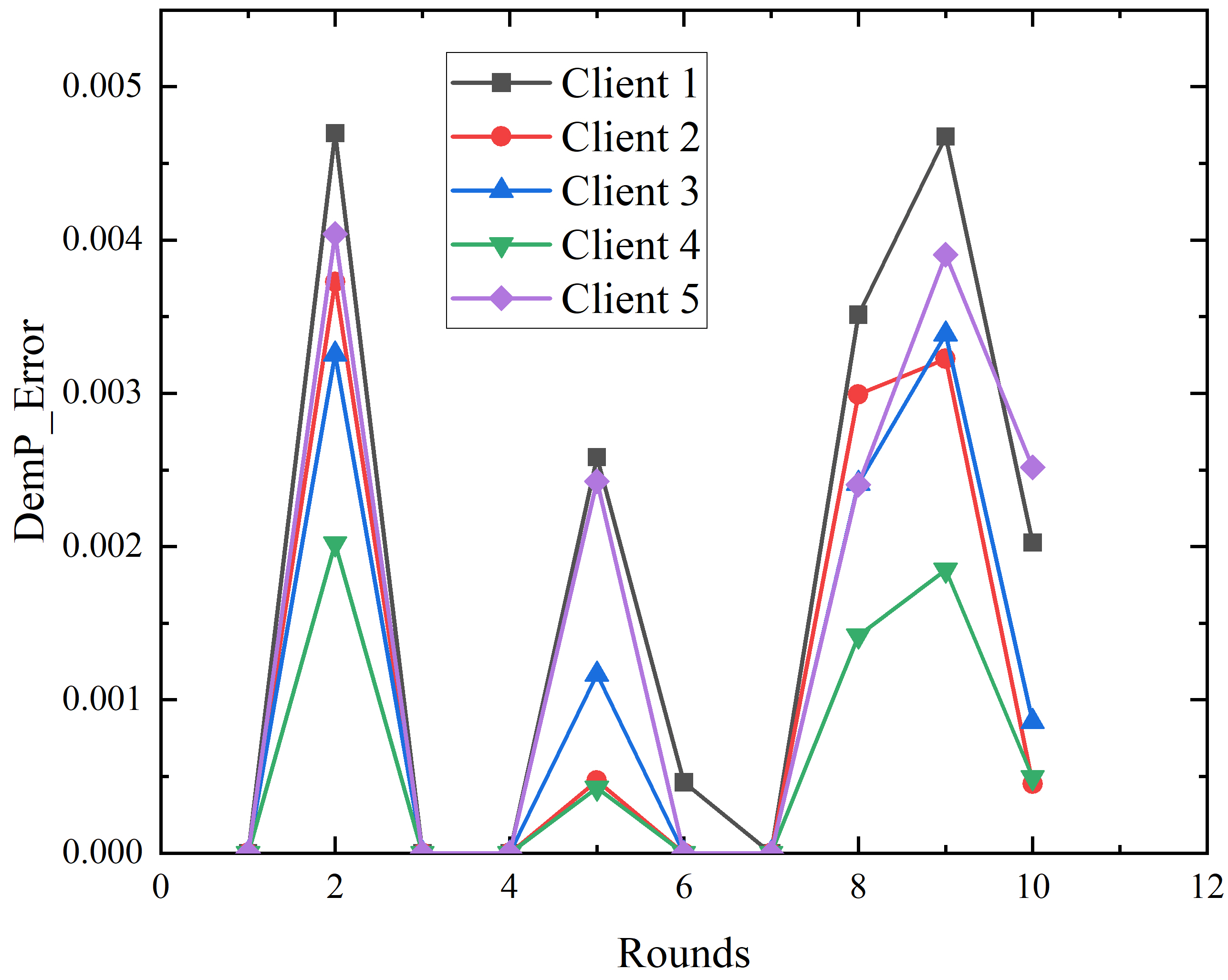}}
        \centerline{(\textit{d}) Demp error with privacy }
        \end{minipage}
    \caption{The EO and DemP error comparison of different clients with privacy and no privacy on \textit{Adult} dataset}
    \label{fig: eo_demp}
\end{figure}
\subsection{Analysis of Privacy and Fairness}
In this section, we analyse the influence of privacy and fairness on the client model. We analyse the fairness metrics of \textit{EO Error} and \textit{DemP Error} to evaluate the error of the training fairness model by adding the privacy ($\sigma = 1$). 
Fig. \ref{fig: eo_demp} (\textit{a})-(\textit{d}) show the \textit{EO Error} and \textit{DemP Error} of different algorithms when each client trains the local fairness model and adds privacy noise. From Fig. \ref{fig: eo_demp} (\textit{a}) and (\textit{c}), the \textit{EO Error} and \textit{DemP Error} without privacy converge to zero.  It can be shown that the client-trained model is fair in both \textit{Demographic Parity} and \textit{Equalized Odds}. However, when privacy is added during federated learning training, the \textit{EO} $l_{EO}$ and \textit{Demp} $l_{Demp}$ loss of the model does not converge, which indicates that adding privacy to the model training process affects the fairness of the model.

\begin{figure}[ht]
    \centering
    \begin{minipage}{1\linewidth}
        \centering
        \centerline{
        \includegraphics[width=1\textwidth]{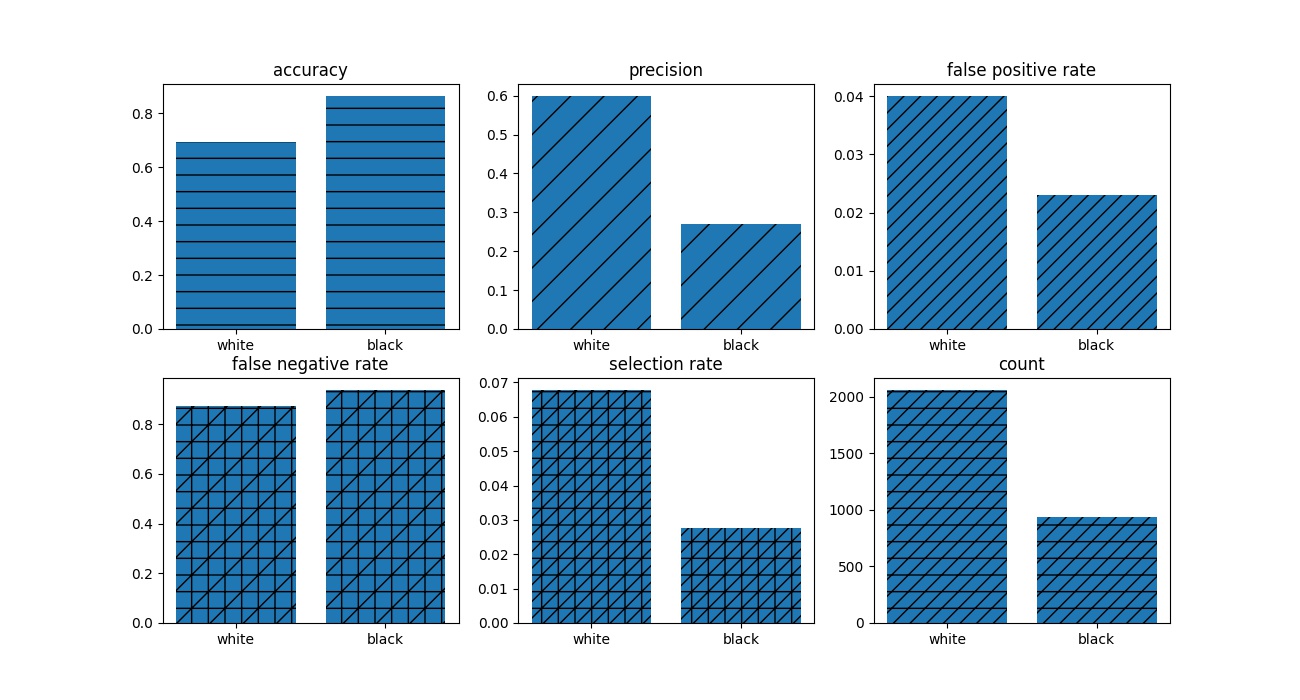}
        }
        \centerline{(\textit{a}) Fair model of client 1 without privacy}
    \end{minipage}
        \begin{minipage}{1\linewidth}
        \centering
        \centerline{
        \includegraphics[width=1\textwidth]{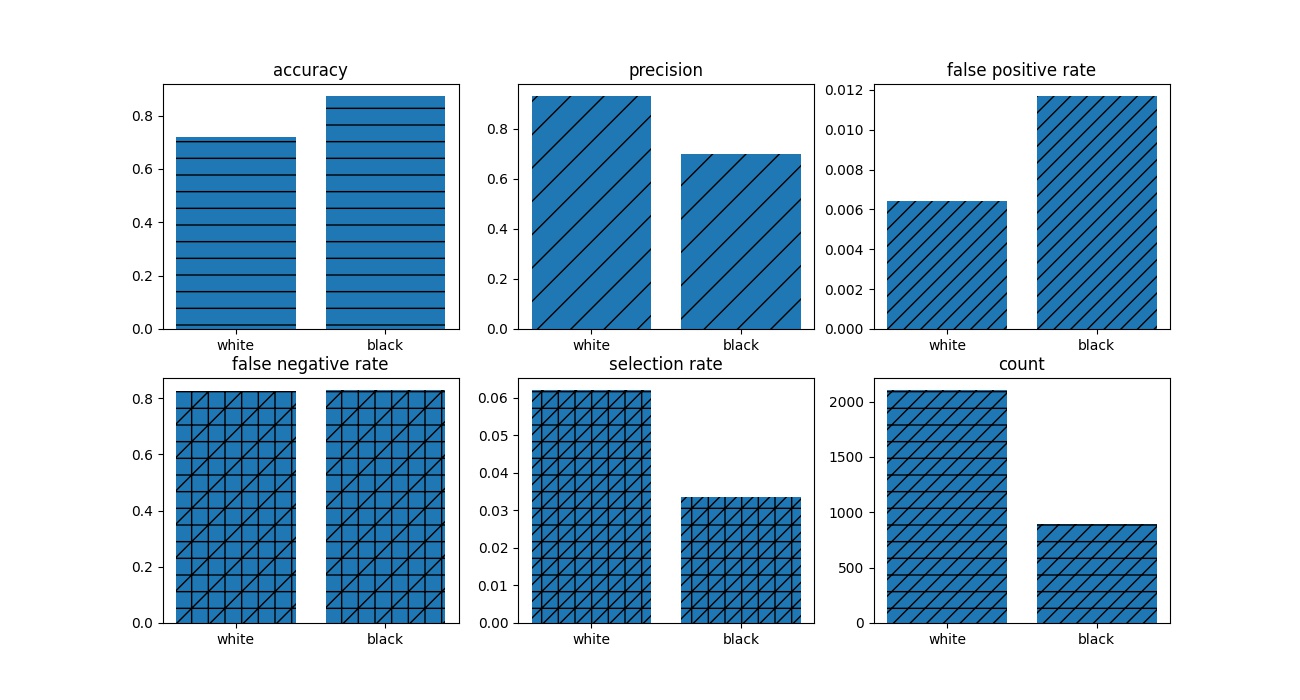}
        }
        \centerline{(\textit{b}) Fair model of client 1 with privacy ($\mathcal{N}(0, 1)$)}
    \end{minipage}
    \caption{The fairness matrics of clients on \textit{Adult} dataset}
    \label{fig:client}
\end{figure}

In Fig. \ref{fig:client}, it is shown the fairness metrics in the client model with privacy and without privacy. In particular, client-side prediction performance is significantly increased by adding noise to the accuracy metric. One of the reasons for this is probably because, with the addition of privacy, the optimizer can jump out of the local optimum in finding the optimal solution.

\section{Conclusion}
In this paper, we research the relationship between fairness and privacy in the FL system. Through the experiment, we found that there is a trade-off between privacy, fairness and accuracy in the FL system. In this paper, we construct the fairness model in clients under the fair metrics constraints, such as \textit{Demographic Parity} (DemP) and \textit{Equqlized Odds} (EOs). In order to protect the fair model privacy, we design a privacy-protecting fairness FL method and we give a private fair algorithm \textit{FedLDP}). In our experiments, we conclude that by adding privacy we can appropriately increase the accuracy of the model while at the same time destroying its fairness. 

\section*{Acknowledgment}
This work was supported in part by the National Natural Science Foundation of China under Grant U20B2048,  62202302.

\bibliographystyle{alpha}
\bibliography{bib}

\newcommand{\etalchar}[1]{$^{#1}$}
\begin{thebibliography}{MMR{\etalchar{+}}17}

\bibitem[ABD{\etalchar{+}}18]{agarwal2018reductions}
Alekh Agarwal, Alina Beygelzimer, Miroslav Dud{\'\i}k, John Langford, and Hanna Wallach.
\newblock A reductions approach to fair classification.
\newblock In {\em International conference on machine learning}, pages 60--69. PMLR, 2018.

\bibitem[AKM20]{awasthi2020equalized}
Pranjal Awasthi, Matth{\"a}us Kleindessner, and Jamie Morgenstern.
\newblock Equalized odds postprocessing under imperfect group information.
\newblock In {\em International conference on artificial intelligence and statistics}, pages 1770--1780. PMLR, 2020.

\bibitem[BBGN19]{balle2019privacy}
Borja Balle, James Bell, Adri{\`a} Gasc{\'o}n, and Kobbi Nissim.
\newblock The privacy blanket of the shuffle model.
\newblock In {\em Advances in Cryptology--CRYPTO 2019: 39th Annual International Cryptology Conference, Santa Barbara, CA, USA, August 18--22, 2019, Proceedings, Part II 39}, pages 638--667. Springer, 2019.

\bibitem[BBGN20]{balle2020private}
Borja Balle, James Bell, Adria Gasc{\'o}n, and Kobbi Nissim.
\newblock Private summation in the multi-message shuffle model.
\newblock In {\em Proceedings of the 2020 ACM SIGSAC Conference on Computer and Communications Security}, pages 657--676, 2020.

\bibitem[BHJ{\etalchar{+}}21]{berk2021fairness}
Richard Berk, Hoda Heidari, Shahin Jabbari, Michael Kearns, and Aaron Roth.
\newblock Fairness in criminal justice risk assessments: The state of the art.
\newblock {\em Sociological Methods \& Research}, 50(1):3--44, 2021.

\bibitem[BPS19]{bagdasaryan2019differential}
Eugene Bagdasaryan, Omid Poursaeed, and Vitaly Shmatikov.
\newblock Differential privacy has disparate impact on model accuracy.
\newblock {\em Advances in neural information processing systems}, 32, 2019.

\bibitem[BWD{\etalchar{+}}22]{bietti2022personalization}
Alberto Bietti, Chen-Yu Wei, Miroslav Dudik, John Langford, and Steven Wu.
\newblock Personalization improves privacy-accuracy tradeoffs in federated learning.
\newblock In {\em International Conference on Machine Learning}, pages 1945--1962. PMLR, 2022.

\bibitem[CCKS22]{chen2022fundamental}
Wei-Ning Chen, Christopher A~Choquette Choo, Peter Kairouz, and Ananda~Theertha Suresh.
\newblock The fundamental price of secure aggregation in differentially private federated learning.
\newblock In {\em International Conference on Machine Learning}, pages 3056--3089. PMLR, 2022.

\bibitem[Cho17]{chouldechova2017fair}
Alexandra Chouldechova.
\newblock Fair prediction with disparate impact: A study of bias in recidivism prediction instruments.
\newblock {\em Big data}, 5(2):153--163, 2017.

\bibitem[CSU{\etalchar{+}}19]{cheu2019distributed}
Albert Cheu, Adam Smith, Jonathan Ullman, David Zeber, and Maxim Zhilyaev.
\newblock Distributed differential privacy via shuffling.
\newblock In {\em Advances in Cryptology--EUROCRYPT 2019: 38th Annual International Conference on the Theory and Applications of Cryptographic Techniques, Darmstadt, Germany, May 19--23, 2019, Proceedings, Part I 38}, pages 375--403. Springer, 2019.

\bibitem[CZZ{\etalchar{+}}23]{chen2023privacy}
Huiqiang Chen, Tianqing Zhu, Tao Zhang, Wanlei Zhou, and Philip~S Yu.
\newblock Privacy and fairness in federated learning: on the perspective of trade-off.
\newblock {\em ACM Computing Surveys}, 2023.

\bibitem[DGK{\etalchar{+}}22]{diana2022multiaccurate}
Emily Diana, Wesley Gill, Michael Kearns, Krishnaram Kenthapadi, Aaron Roth, and Saeed Sharifi-Malvajerdi.
\newblock Multiaccurate proxies for downstream fairness.
\newblock In {\em Proceedings of the 2022 ACM Conference on Fairness, Accountability, and Transparency}, pages 1207--1239, 2022.

\bibitem[DLC{\etalchar{+}}20]{duan2020self}
Moming Duan, Duo Liu, Xianzhang Chen, Renping Liu, Yujuan Tan, and Liang Liang.
\newblock Self-balancing federated learning with global imbalanced data in mobile systems.
\newblock {\em IEEE Transactions on Parallel and Distributed Systems}, 32(1):59--71, 2020.

\bibitem[DR{\etalchar{+}}14]{dwork2014algorithmic}
Cynthia Dwork, Aaron Roth, et~al.
\newblock The algorithmic foundations of differential privacy.
\newblock {\em Foundations and Trends{\textregistered} in Theoretical Computer Science}, 9(3--4):211--407, 2014.

\bibitem[EFM{\etalchar{+}}19]{erlingsson2019amplification}
{\'U}lfar Erlingsson, Vitaly Feldman, Ilya Mironov, Ananth Raghunathan, Kunal Talwar, and Abhradeep Thakurta.
\newblock Amplification by shuffling: From local to central differential privacy via anonymity.
\newblock In {\em Proceedings of the Thirtieth Annual ACM-SIAM Symposium on Discrete Algorithms}, pages 2468--2479. SIAM, 2019.

\bibitem[EGLC22]{esipova2022disparate}
Maria~S Esipova, Atiyeh~Ashari Ghomi, Yaqiao Luo, and Jesse~C Cresswell.
\newblock Disparate impact in differential privacy from gradient misalignment.
\newblock {\em arXiv preprint arXiv:2206.07737}, 2022.

\bibitem[FMST20]{farrand2020neither}
Tom Farrand, Fatemehsadat Mireshghallah, Sahib Singh, and Andrew Trask.
\newblock Neither private nor fair: Impact of data imbalance on utility and fairness in differential privacy.
\newblock In {\em Proceedings of the 2020 workshop on privacy-preserving machine learning in practice}, pages 15--19, 2020.

\bibitem[GDD{\etalchar{+}}21]{girgis2021shuffled}
Antonious Girgis, Deepesh Data, Suhas Diggavi, Peter Kairouz, and Ananda~Theertha Suresh.
\newblock Shuffled model of differential privacy in federated learning.
\newblock In {\em International Conference on Artificial Intelligence and Statistics}, pages 2521--2529. PMLR, 2021.

\bibitem[GGK{\etalchar{+}}21]{ghazi2021power}
Badih Ghazi, Noah Golowich, Ravi Kumar, Rasmus Pagh, and Ameya Velingker.
\newblock On the power of multiple anonymous messages: Frequency estimation and selection in the shuffle model of differential privacy.
\newblock In {\em Annual International Conference on the Theory and Applications of Cryptographic Techniques}, pages 463--488. Springer, 2021.

\bibitem[GKN17]{geyer2017differentially}
Robin~C Geyer, Tassilo Klein, and Moin Nabi.
\newblock Differentially private federated learning: A client level perspective.
\newblock {\em arXiv preprint arXiv:1712.07557}, 2017.

\bibitem[GMS{\etalchar{+}}23]{gehlhar2023safefl}
Till Gehlhar, Felix Marx, Thomas Schneider, Ajith Suresh, Tobias Wehrle, and Hossein Yalame.
\newblock Safefl: Mpc-friendly framework for private and robust federated learning.
\newblock {\em Cryptology ePrint Archive}, 2023.

\bibitem[GODC22]{ganev2022robin}
Georgi Ganev, Bristena Oprisanu, and Emiliano De~Cristofaro.
\newblock Robin hood and matthew effects: Differential privacy has disparate impact on synthetic data.
\newblock In {\em International Conference on Machine Learning}, pages 6944--6959. PMLR, 2022.

\bibitem[{Hao}21]{hao2021towards}
{Hao, Weituo and El-Khamy, Mostafa and Lee, Jungwon and Zhang, Jianyi and Liang, Kevin J and Chen, Changyou and Duke, Lawrence Carin}.
\newblock Towards fair federated learning with zero-shot data augmentation.
\newblock In {\em Proceedings of the IEEE/CVF Conference on Computer Vision and Pattern Recognition}, pages 3310--3319, 2021.

\bibitem[HPS16]{hardt2016equality}
Moritz Hardt, Eric Price, and Nati Srebro.
\newblock Equality of opportunity in supervised learning.
\newblock {\em Advances in neural information processing systems}, 29, 2016.

\bibitem[JKM{\etalchar{+}}19]{jagielski2019differentially}
Matthew Jagielski, Michael Kearns, Jieming Mao, Alina Oprea, Aaron Roth, Saeed Sharifi-Malvajerdi, and Jonathan Ullman.
\newblock Differentially private fair learning.
\newblock In {\em International Conference on Machine Learning}, pages 3000--3008. PMLR, 2019.

\bibitem[JOK{\etalchar{+}}18]{jeong2018communication}
Eunjeong Jeong, Seungeun Oh, Hyesung Kim, Jihong Park, Mehdi Bennis, and Seong-Lyun Kim.
\newblock Communication-efficient on-device machine learning: Federated distillation and augmentation under non-iid private data.
\newblock {\em arXiv preprint arXiv:1811.11479}, 2018.

\bibitem[KGK{\etalchar{+}}18]{kilbertus2018blind}
Niki Kilbertus, Adri{\`a} Gasc{\'o}n, Matt Kusner, Michael Veale, Krishna Gummadi, and Adrian Weller.
\newblock Blind justice: Fairness with encrypted sensitive attributes.
\newblock In {\em International Conference on Machine Learning}, pages 2630--2639. PMLR, 2018.

\bibitem[KKM{\etalchar{+}}20]{karimireddy2020scaffold}
Sai~Praneeth Karimireddy, Satyen Kale, Mehryar Mohri, Sashank Reddi, Sebastian Stich, and Ananda~Theertha Suresh.
\newblock Scaffold: Stochastic controlled averaging for federated learning.
\newblock In {\em International conference on machine learning}, pages 5132--5143. PMLR, 2020.

\bibitem[KMA{\etalchar{+}}21]{kairouz2021advances}
Peter Kairouz, H~Brendan McMahan, Brendan Avent, Aur{\'e}lien Bellet, Mehdi Bennis, Arjun~Nitin Bhagoji, Kallista Bonawitz, Zachary Charles, Graham Cormode, Rachel Cummings, et~al.
\newblock Advances and open problems in federated learning.
\newblock {\em Foundations and Trends{\textregistered} in Machine Learning}, 14(1--2):1--210, 2021.

\bibitem[LGR23a]{lowy2023private}
Andrew Lowy, Ali Ghafelebashi, and Meisam Razaviyayn.
\newblock Private non-convex federated learning without a trusted server.
\newblock In {\em International Conference on Artificial Intelligence and Statistics}, pages 5749--5786. PMLR, 2023.

\bibitem[LGR23b]{lowy2023stochastic}
Andrew Lowy, Devansh Gupta, and Meisam Razaviyayn.
\newblock Stochastic differentially private and fair learning.
\newblock In {\em Workshop on Algorithmic Fairness through the Lens of Causality and Privacy}, pages 86--119. PMLR, 2023.

\bibitem[LLF{\etalchar{+}}23]{li2023privacy}
Xiaochen Li, Weiran Liu, Hanwen Feng, Kunzhe Huang, Yuke Hu, Jinfei Liu, Kui Ren, and Zhan Qin.
\newblock Privacy enhancement via dummy points in the shuffle model.
\newblock {\em IEEE Transactions on Dependable and Secure Computing}, 2023.

\bibitem[LSBS19]{li2019fair}
Tian Li, Maziar Sanjabi, Ahmad Beirami, and Virginia Smith.
\newblock Fair resource allocation in federated learning.
\newblock {\em arXiv preprint arXiv:1905.10497}, 2019.

\bibitem[LZMV19]{lamy2019noise}
Alex Lamy, Ziyuan Zhong, Aditya~K Menon, and Nakul Verma.
\newblock Noise-tolerant fair classification.
\newblock {\em Advances in neural information processing systems}, 32, 2019.

\bibitem[MBS20]{martinez2020minimax}
Natalia Martinez, Martin Bertran, and Guillermo Sapiro.
\newblock Minimax pareto fairness: A multi objective perspective.
\newblock In {\em International Conference on Machine Learning}, pages 6755--6764. PMLR, 2020.

\bibitem[MMR{\etalchar{+}}17]{mcmahan2017communication}
Brendan McMahan, Eider Moore, Daniel Ramage, Seth Hampson, and Blaise~Aguera y~Arcas.
\newblock Communication-efficient learning of deep networks from decentralized data.
\newblock In {\em Artificial intelligence and statistics}, pages 1273--1282. PMLR, 2017.

\bibitem[MOS20]{mozannar2020fair}
Hussein Mozannar, Mesrob Ohannessian, and Nathan Srebro.
\newblock Fair learning with private demographic data.
\newblock In {\em International Conference on Machine Learning}, pages 7066--7075. PMLR, 2020.

\bibitem[PG20]{Adult}
Manisha Padala and Sujit Gujar.
\newblock Fnnc: Achieving fairness through neural networks.
\newblock In {\em Proceedings of the Twenty-Ninth International Joint Conference on Artificial Intelligence,$\{$IJCAI-20$\}$, International Joint Conferences on Artificial Intelligence Organization}, 2020.

\bibitem[PMK{\etalchar{+}}20]{pujol2020fair}
David Pujol, Ryan McKenna, Satya Kuppam, Michael Hay, Ashwin Machanavajjhala, and Gerome Miklau.
\newblock Fair decision making using privacy-protected data.
\newblock In {\em Proceedings of the 2020 Conference on Fairness, Accountability, and Transparency}, pages 189--199, 2020.

\bibitem[RSL{\etalchar{+}}08]{raskhodnikova2008can}
Sofya Raskhodnikova, Adam Smith, Homin~K Lee, Kobbi Nissim, and Shiva~Prasad Kasiviswanathan.
\newblock What can we learn privately.
\newblock In {\em Proceedings of the 54th Annual Symposium on Foundations of Computer Science}, pages 531--540, 2008.

\bibitem[SLS{\etalchar{+}}23]{shao2023survey}
Jiawei Shao, Zijian Li, Wenqiang Sun, Tailin Zhou, Yuchang Sun, Lumin Liu, Zehong Lin, and Jun Zhang.
\newblock A survey of what to share in federated learning: Perspectives on model utility, privacy leakage, and communication efficiency.
\newblock {\em arXiv preprint arXiv:2307.10655}, 2023.

\bibitem[SMS22]{scheliga2022precode}
Daniel Scheliga, Patrick M{\"a}der, and Marco Seeland.
\newblock Precode-a generic model extension to prevent deep gradient leakage.
\newblock In {\em Proceedings of the IEEE/CVF Winter Conference on Applications of Computer Vision}, pages 1849--1858, 2022.

\bibitem[TFVH21]{tran2021differentially}
Cuong Tran, Ferdinando Fioretto, and Pascal Van~Hentenryck.
\newblock Differentially private and fair deep learning: A lagrangian dual approach.
\newblock In {\em Proceedings of the AAAI Conference on Artificial Intelligence}, volume~35, pages 9932--9939, 2021.

\bibitem[WGN{\etalchar{+}}20]{wang2020robust}
Serena Wang, Wenshuo Guo, Harikrishna Narasimhan, Andrew Cotter, Maya Gupta, and Michael Jordan.
\newblock Robust optimization for fairness with noisy protected groups.
\newblock {\em Advances in neural information processing systems}, 33:5190--5203, 2020.

\bibitem[WKL{\etalchar{+}}21]{wu2021fedcg}
Yuezhou Wu, Yan Kang, Jiahuan Luo, Yuanqin He, and Qiang Yang.
\newblock Fedcg: Leverage conditional gan for protecting privacy and maintaining competitive performance in federated learning.
\newblock {\em arXiv preprint arXiv:2111.08211}, 2021.

\bibitem[WKNL20]{wang2020optimizing}
Hao Wang, Zakhary Kaplan, Di~Niu, and Baochun Li.
\newblock Optimizing federated learning on non-iid data with reinforcement learning.
\newblock In {\em IEEE INFOCOM 2020-IEEE Conference on Computer Communications}, pages 1698--1707. IEEE, 2020.

\bibitem[WLD{\etalchar{+}}20]{wei2020federated}
Kang Wei, Jun Li, Ming Ding, Chuan Ma, Howard~H Yang, Farhad Farokhi, Shi Jin, Tony~QS Quek, and H~Vincent Poor.
\newblock Federated learning with differential privacy: Algorithms and performance analysis.
\newblock {\em IEEE Transactions on Information Forensics and Security}, 15:3454--3469, 2020.

\bibitem[XBJ21]{xu2021privacy}
Runhua Xu, Nathalie Baracaldo, and James Joshi.
\newblock Privacy-preserving machine learning: Methods, challenges and directions.
\newblock {\em arXiv preprint arXiv:2108.04417}, 2021.

\bibitem[YLL{\etalchar{+}}20]{yu2020fairness}
Han Yu, Zelei Liu, Yang Liu, Tianjian Chen, Mingshu Cong, Xi~Weng, Dusit Niyato, and Qiang Yang.
\newblock A fairness-aware incentive scheme for federated learning.
\newblock In {\em Proceedings of the AAAI/ACM Conference on AI, Ethics, and Society}, pages 393--399, 2020.

\bibitem[ZVRG17]{zafar2017fairness}
Muhammad~Bilal Zafar, Isabel Valera, Manuel~Gomez Rogriguez, and Krishna~P Gummadi.
\newblock Fairness constraints: Mechanisms for fair classification.
\newblock In {\em Artificial intelligence and statistics}, pages 962--970. PMLR, 2017.

\bibitem[ZXW{\etalchar{+}}22]{zhou2022multi}
Zan Zhou, Changqiao Xu, Mingze Wang, Xiaohui Kuang, Yirong Zhuang, and Shui Yu.
\newblock A multi-shuffler framework to establish mutual confidence for secure federated learning.
\newblock {\em IEEE Transactions on Dependable and Secure Computing}, 2022.

\end{thebibliography}
\end{document}